\documentclass[conference]{IEEEtran}
\IEEEoverridecommandlockouts
\usepackage{cite}
\usepackage{amsmath,amssymb,amsfonts}
\usepackage{algorithmicx}
\usepackage{graphicx}
\usepackage{textcomp}
\usepackage{multirow} 
\usepackage{booktabs}
\usepackage{url}
\usepackage{booktabs}       

\usepackage{amsfonts,amsmath}       
\usepackage{nicefrac}       
\usepackage{microtype}      

\usepackage{multirow}       
\usepackage{graphics}
\usepackage{tabularx}
\usepackage{mathrsfs}
\usepackage{makecell}
\usepackage{caption}
\usepackage{wrapfig}
\usepackage{enumitem}
\usepackage{subcaption}
\usepackage{bbding}
\usepackage{algpseudocode}
\usepackage{amsfonts}       
\usepackage[colorlinks,citecolor=blue,linkcolor=red,bookmarks=false]{hyperref}
\usepackage{color, colortbl}
\usepackage{xcolor}
\definecolor{Gray}{gray}{0.95}
\definecolor{orange}{rgb}{0.9,0.5,0}
\definecolor{LightCyan}{rgb}{0.88,1,1}

\def\BibTeX{{\rm B\kern-.05em{\sc i\kern-.025em b}\kern-.08em
    T\kern-.1667em\lower.7ex\hbox{E}\kern-.125emX}}
\begin{document}

\title{Fraesormer: Learning Adaptive Sparse Transformer for Efficient Food Recognition}

\author{
	\IEEEauthorblockN{
		Shun Zou\IEEEauthorrefmark{1}\thanks{$^{\dagger}$Corresponding author.}, 
		Yi Zou\IEEEauthorrefmark{2}, 
        Mingya Zhang\IEEEauthorrefmark{3},
		Shipeng Luo\IEEEauthorrefmark{4},
        Zhihao Chen\IEEEauthorrefmark{5},
        Guangwei Gao\IEEEauthorrefmark{6}$^{\dagger}$} 
        
        \IEEEauthorblockA{%
      \begin{tabular}{cccc}
        \IEEEauthorrefmark{1}Nanjing Agricultural University & \IEEEauthorrefmark{2}Xiangtan University &
        \IEEEauthorrefmark{3}Nanjing University &
        \IEEEauthorrefmark{4}Northeast Forestry University 
      \end{tabular}%
    }
    
    \IEEEauthorblockA{%
  \begin{tabular}{cc}
    \IEEEauthorrefmark{5}Beijing Information Science and Technology University &
    \IEEEauthorrefmark{6}Nanjing University of Posts and Telecommunications
  \end{tabular}%
}

	\IEEEauthorblockA{
 zs@stu.njau.edu.cn, csgwgao@njupt.edu.cn}

}

\DeclareRobustCommand*{\IEEEauthorrefmark}[1]{%
    \raisebox{0pt}[0pt][0pt]{\textsuperscript{\footnotesize\ensuremath{#1}}}}

\maketitle

\begin{abstract}
In recent years, Transformer has witnessed significant progress in food recognition. However, most existing approaches still face two critical challenges in lightweight food recognition: (1) the quadratic complexity and redundant feature representation from interactions with irrelevant tokens; (2) static feature recognition and single-scale representation, which overlook the unstructured, non-fixed nature of food images and the need for multi-scale features. To address these, we propose an adaptive and efficient sparse Transformer architecture (Fraesormer) with two core designs: Adaptive Top-k Sparse Partial Attention (ATK-SPA) and Hierarchical Scale-Sensitive Feature Gating Network (HSSFGN). ATK-SPA uses a learnable Gated Dynamic Top-K Operator (GDTKO) to retain critical attention scores, filtering low query-key matches that hinder feature aggregation. It also introduces a partial channel mechanism to reduce redundancy and promote expert information flow, enabling local-global collaborative modeling. HSSFGN employs gating mechanism to achieve multi-scale feature representation, enhancing contextual semantic information. Extensive experiments show that Fraesormer outperforms state-of-the-art methods. code is available at \href{https://zs1314.github.io/Fraesormer}{https://zs1314.github.io/Fraesormer}.
\end{abstract}


\section{Introduction}
Food plays a central role in our daily lives, and food computing can assist in areas such as diet, health, and the food industry \cite{mavani2022application,boswell2018training}. With the rise of deep learning, food computing has gained increasing attention and undergone significant development in computer vision and multimedia \cite{rostami2022world,shao2023vision}. Furthermore, as the primary goal of food computing systems is to help individuals manage their diet and health while enhancing their daily activities, it is essential to develop an efficient system specifically designed for food image recognition on edge devices \cite{yang2024lightweight}.


Many learning-based methods have recently utilized various CNN architectures for food recognition \cite{wang2022ingredient,kiourt2020deep}. Although CNNs achieve good efficiency and generalization due to their local connections and translation invariance, the receptive field of the convolution operator is limited, preventing them from capturing global dependencies and modeling long-range pixel relationships. In food recognition, food images often contain multiple ingredients that are sparsely distributed, 
\begin{figure}[ht]
	\centering
	\includegraphics[width=0.5\textwidth]{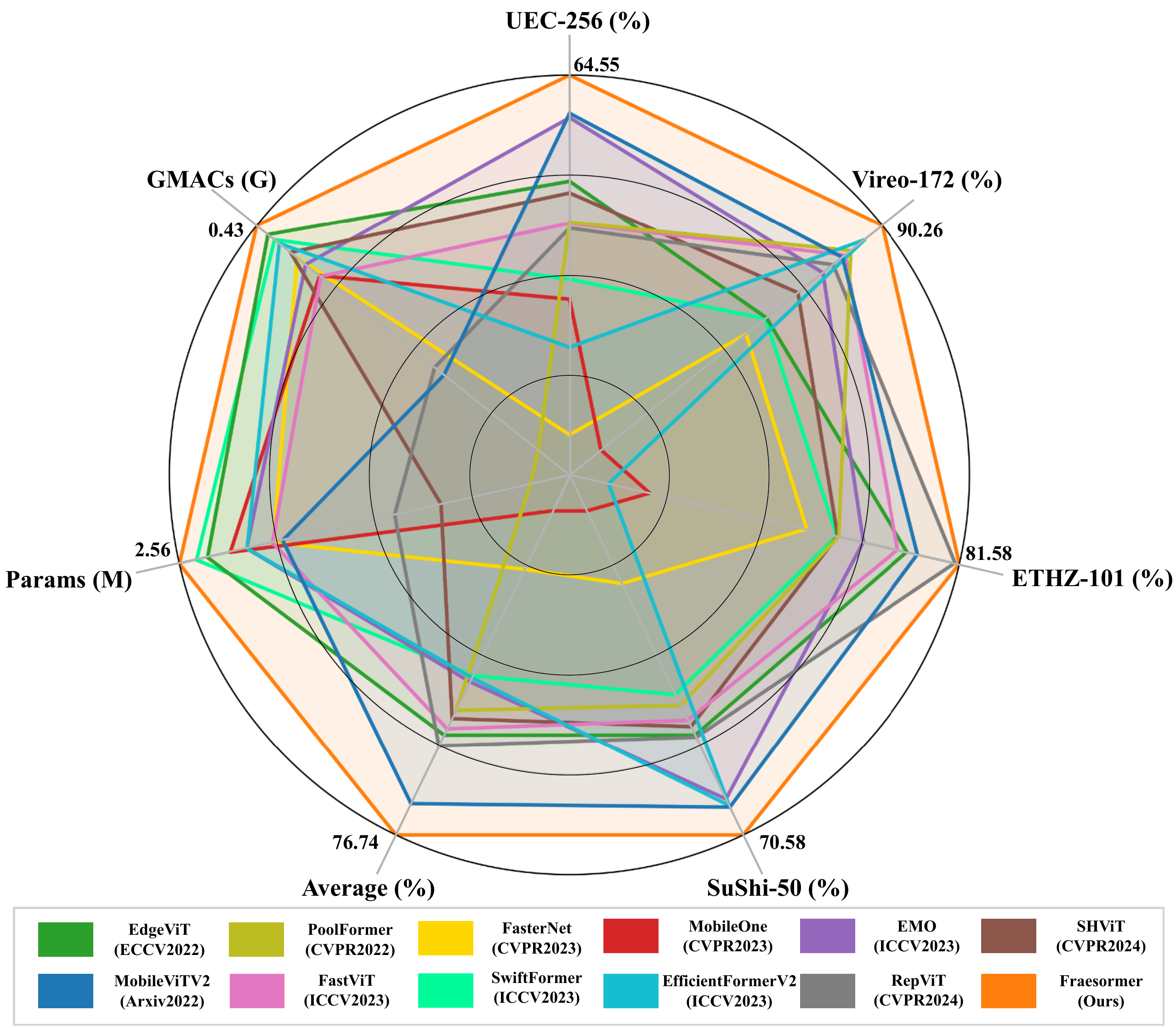} 
	\vspace{-5mm}
	\caption{A seven-dimensional radar map of the Top-1 Acc of ETHZ-101 \cite{food-101}, Vireo-172 \cite{food-172}, UEC-256 \cite{food-256}, SuShi-50 \cite{qiu2019mining}, along with Average Acc, Params, and GMACs.}
	\label{compare}
	\vspace{-6mm}
\end{figure}
 adding complexity and requiring fine-grained analysis \cite{rodenas2022learning}. Therefore, there is a greater demand for neural networks capable of extracting long-range relationships \cite{yang2024lightweight}. To address this challenge, the self-attention mechanism offers a more powerful alternative \cite{dosovitskiy2020vit}.
Self-attention is the core component of Transformers, showing state-of-the-art performance in 
computer vision \cite{li2022rethinking,vasufastvit2023}. Researchers have also applied Transformers to food recognition \cite{zhu2020food,bianco2023food}, achieving promising results due to their ability to extract non-local information and model global states effectively. However, while Transformers excel at capturing long-range pixel interactions, their complexity increases quadratically with spatial resolution, resulting in significant computational burdens. This poses a considerable challenge for developing a lightweight recognition system. To date, research on lightweight Transformer-based models specifically tailored for food recognition remains limited.
\begin{figure*}[!ht]
	\centering
	\includegraphics[width=\linewidth]{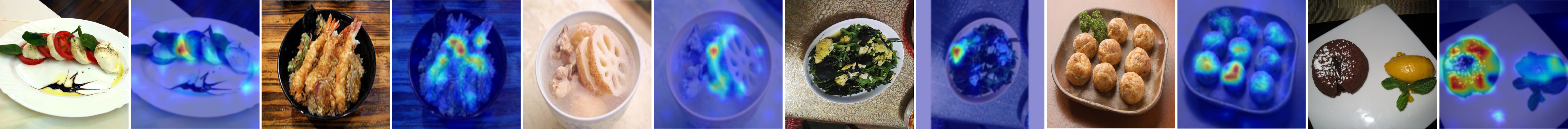} 
	\vspace{-5mm}
	\caption{Visualization of the impact of each spatial location on the final prediction of the DeiT-S model \cite{Chefer_2021_CVPR,touvron2021training}. The results show that the final prediction of the vision transformer is primarily based on the most influential tokens, indicating that a large portion of tokens can be removed without affecting performance.}
	\label{vis}
	\vspace{-6mm}
\end{figure*}

To address these issues, we re-examine the challenges associated with lightweight food recognition: (1) Although some studies have attempted to simplify the attention mechanism \cite{vasufastvit2023,mehta2022separable}, the number of parameters and computational costs remain high. Moreover, we observe that standard Transformers aggregate features based on all attention relationships between queries and keys; however, the tokens in the keys are not always relevant to those in the queries \cite{wang2022kvt} (see Fig. \ref{vis}). Using irrelevant tokens to estimate self-attention values leads to computational overhead, information redundancy, and noisy interactions during feature aggregation; (2) Traditional visual recognition methods struggle to handle the unstructured and variable nature of food images, necessitating an adaptive feature extraction paradigm \cite{10471331}. For example, the appearance of food can change significantly during cooking; the visual features of raw fish slices differ greatly from those of cooked fish fillets. Traditional feature extraction methods cannot easily adapt to such dynamic changes. Moreover, some dishes can be easily confused due to similarities in cooking methods. For instance,  ``beef curry" and  ``braised beef with potatoes" may be misclassified because of similar preparation techniques (``curry braise" and  ``potato braise"). Therefore, an adaptive feature extraction mechanism is more effective in addressing challenges such as changes in ingredient morphology, regional differences, and similar cooking styles; (3) Multi-scale features are crucial for accurately recognizing food semantics. In a French dish, both the main ingredient and small components contribute to its overall characteristics. For example, in a foie gras dish, while the large foie gras portion is the main element, the small fruits and herbs are also key to identification. If the model focuses solely on the main ingredient, it may fail to capture complete semantic information, leading to misinterpretation.

To address the above challenges, we propose a \textbf{f}ood \textbf{r}ecognition-focused \textbf{a}daptive \textbf{e}fficient \textbf{s}parse Transf\textbf{ormer} network, named \textbf{Fraesormer}. Specifically, the key component of this framework is the Fraesormer Block, which includes Adaptive Top-k Sparse Partial Attention (ATK-SPA) for extracting the most significant attention values and the Hierarchical Scale-Sensitive Feature Gating Network (HSSFGN) for exploring multi-scale contextual local information. More precisely, ATK-SPA does not compute dense attention matrices using all tokens. Instead, it adapts the value of $k$ based on the input features by using a novel and efficient Gated Dynamic Top-K Operator (GDTKO). It selects the top-k tokens with the highest attention values to filter out irrelevant noise, fully leveraging the network's sparsity while ensuring a dynamic feature extraction paradigm. Additionally, we have combined a parallel partial channel mechanism with the attention mechanism, effectively addressing channel redundancy while enhancing local information. This approach utilizes two sets of complementary features to collaboratively model both local and non-local features. Furthermore, the HSSFGN we designed further exploits cross-level multi-scale information for improved feature aggregation. Finally, comprehensive experiments show that Fraesormer achieves the better performance-parameter trade-off, making it a true “Heptagon Warrior", as shown in Fig. \ref{compare}. 
\begin{figure*}[ht]
	\centering
	\includegraphics[width=\linewidth]{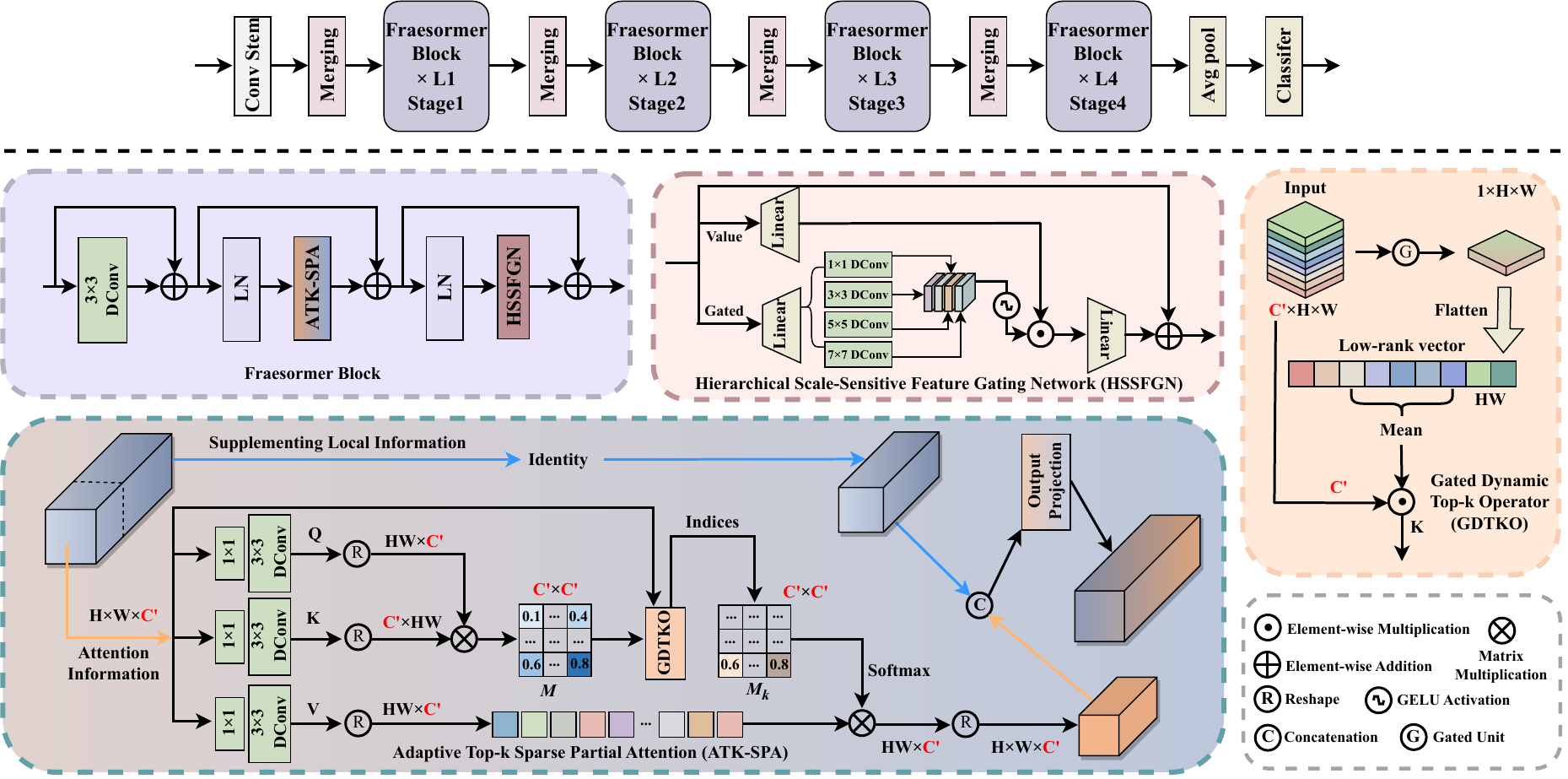} 
	\vspace{-5mm}
	\caption{The overall architecture of the proposed Fraesormer.}
	\label{overview}
	\vspace{-6mm}
\end{figure*}
The main contributions of this work are summarized as follows:
\begin{itemize}
    \item We propose an efficient adaptive sparse Transformer architecture that promotes the flow of the most valuable information across layers, effectively addressing the challenges faced in lightweight food recognition. 
    \item We introduce Adaptive Top-k Sparse Partial Attention, which adaptively retains the most useful features, reducing redundant information and alleviating computational burden. Additionally, we design a novel hybrid-scale feedforward network based on the gating mechanism 
    paradigm to efficiently explore cross-level multi-scale representations.
    \item We conducted comprehensive experiments on four widely used authoritative food datasets, and the results demonstrate that our method outperforms state-of-the-art approaches. Furthermore, we provide extensive ablation studies to highlight the contributions of the design.
\end{itemize}

\section{Method}
\subsection{Overall Pipeline}
The overall pipeline of our Fraesormer is illustrated in Fig. \ref {overview}. In line with previous general visual backbones \cite{fan2023rmt}, the Fraesormer is divided into four stages, each consisting of several Fraesormer Blocks. Additionally, each stage is preceded by a merging layer for spatial downsampling and channel expansion. We apply a global average pooling layer to the final output and send it to a linear classification head. Similar to earlier backbone networks \cite{guo2022cmt,zhu2023biformer}, we incorporate CPE \cite{chu2023CPVT} into our model. In each block, we have developed ATK-SPA, which differs from standard self-attention in Transformers, to achieve feature sparsity and execute the feature aggregation process more efficiently. We also introduce a parallel partial channel mechanism to reduce redundant information and enhance the retention of local features, facilitating the expansion of channel information. Furthermore, we integrate HSSFGN within the Fraesormer Block to extract rich local scale information across different levels while filtering out hidden redundant information in the feature maps. Formally, given the input \( X_{l-1} \) of the \( l-1 \) block, the processing of the Fraesormer Block is defined as follows:
\vspace{-2mm}
\begin{equation}
    X_{l}=X_{l-1}+DW(X_{l-1}),
    \vspace{-2mm}
\end{equation}
\begin{equation}
X_{l}^{\prime}=X_{l}+ATK\textbf{-}SPA(LN(X_{l})),
   \vspace{-2mm}
\end{equation}
\begin{equation}
    X_{l}'^{\prime}=X_{l}'+HSSFGN(LN(X_{l}')),
    \vspace{-2mm}
\end{equation}
where \( DW(\cdot) \) represents the depth-wise convolution operation, \( LN(\cdot) \) denotes layer normalization, and \( X_l, X_{l}', X_{l}'' \) are the outputs of DW, ATK-SPA, and HSSFGN, respectively.

\subsection{Adaptive Top-k Sparse Partial Attention}
Our ATK-SPA is motivated by two main objectives: (1) to extract only the most critical information during attention calculation, thereby eliminating the influence of noise tokens while minimizing redundant information to reduce computational burden; and (2) to achieve adaptive aggregation and dynamic filtering of image features, proposing a feature extraction strategy that flexibly adapts to the complexity of food images and the irregularity of visual patterns, effectively addressing diverse appearances and semantic content.

Fig. \ref{overview} illustrates the architecture of ATK-SPA. Specifically, given the input normalized feature \( F \in\mathbb{R}^{H \times W \times C} \), we selectively choose a subset of channels for spatial feature aggregation while keeping the remaining channels unchanged, resulting in \( X_{att} \in \mathbb{R}^{H \times W \times C'} \) and \( X_{sup} \in \mathbb{R}^{H \times W \times C''} \). Next, we apply a \( 1 \times 1 \) point-wise convolution on \( X_{att} \) to aggregate local information across pixel channels, and use \( 3 \times 3 \) depth-wise convolutions to enhance spatial context across channels, generating the matrices for query \( Q \), key \( K \), and value \( V \). We then perform a dot-product operation on the reshaped \( Q \) and \( K \) to produce a dense attention matrix \( M \in \mathbb{R}^{C' \times C'} \). In contrast to calculating a spatial matrix with dimensions \( HW \times HW \), measuring channel similarity helps reduce memory consumption, thereby enabling efficient inference \cite{Zamir2021Restormer}. We re-examine the standard dense self-attention mechanism (SDSA) \cite{dosovitskiy2020vit} used in most previous work, which computes the attention map for all query-key pairs:
\vspace{-2mm}
\begin{equation}
    SDSA=Softmax(\frac{QK^{T}}{\sqrt{d} }+ B )V.
\vspace{-3mm}
\end{equation}
Here, \( \sqrt{d} \) represents an optional temperature coefficient used to control the size of the dot product between \( Q \) and \( K \) before applying softmax function, and \( B \) denotes a learnable relative position bias. However, not all tokens in the queries are closely related to their corresponding tokens in the keys, so calculating and utilizing the similarities of all query-key pairs can lead to noisy interactions and information redundancy. To address this issue, we perform adaptive selection on the dense attention matrix \( M \), retaining only the top \( k \) attention scores with the highest contributions. This approach aims to preserve the most useful information in the attention matrix while filtering out the majority of noise interference. The selection of \( k \) is critical; instead of fixing \( k \), we adaptively adjust it based on the input using a novel, simple yet efficient Gated Dynamic Top-K Operator. This enables a dynamic selection process that transitions the attention matrix from dense to sparse (which will be elaborated upon in the next subsection). For example, with \( k = 2/3 \), we retain only the top \( 2/3 \) of the highest attention scores while masking the remaining elements to zero. Specifically, we generate a binary mask matrix:
\vspace{-2mm}
\begin{equation}
	\left[{M}_k\right]_{i j}= \begin{cases}1 & M_{i j} \in indices_{k}
	\\ 0 & \text { otherwise}\end{cases},
\vspace{-2mm}
\end{equation}
where \( \text{indices}_k \) represents the indices of the top \( k \) highest values. Formally, ATK-SPA can be described as:
\vspace{-2mm}
\begin{equation}
    X_{att},X_{sup}=Split(X),
    \vspace{-2mm}
\end{equation}
\begin{equation}
    ATK\text{-}SPA=Softmax(T_{k}(\frac{QK^{T}}{\sqrt{d} } +B)) V,
    \vspace{-2mm}
\end{equation}
\begin{equation}
    \hat{X_{att}}=ATK\text{-}SPA (X_{att}W_{d}^{Q}W_{p}^{Q},X_{att}W_{d}^{K}W_{p}^{K},X_{att}W_{d}^{V}W_{p}^{V}),
    \vspace{-1mm}
\end{equation}
\begin{equation}
    X'=Concat(\hat{X_{att}},X_{sup} )W_{p}^{O},
    \vspace{-2mm}
\end{equation}
\begin{table}[htp]
\begin{center}
  \caption{Comparison of quantitative results on four food datasets. \textbf{Bold} values indicate the best result. $\uparrow$ means higher is better, $\downarrow$ means lower is better. }
  \label{classification}
  \vspace{-0.2cm}
  \resizebox{1\linewidth}{!}{
    \begin{tabular}{l|cccccccc}
    \toprule[1.8pt]
     Model & \makecell{Params \\ (M)$\downarrow$} & \makecell{MACs \\ (G)$\downarrow$} & \makecell{UEC-256 \cite{food-256}  \\Top-1(\%)$\uparrow$}& \makecell{Vireo-172 \cite{food-172}  \\Top-1(\%)$\uparrow$} & \makecell{ETHZ-101 \cite{food-101}  \\Top-1(\%)$\uparrow$} & \makecell{SuShi-50 \cite{qiu2019mining}  \\Top-1(\%)$\uparrow$}  & \makecell{Average  \\Top-1(\%)$\uparrow$}  & \makecell{Year}    \\
    \midrule[1.5pt]
    EMO-1M \cite{emo} & 1.36 & 0.26 & 62.882 & 87.560 & 80.267 & 54.096 & 71.201 & ICCV2023\\
    RepViT-M0-6 \cite{wang2024repvit} & 2.49 & 0.40 & 62.018 & 87.693 & 79.228 & 55.060 & 71.000 & CVPR2024\\
    MobileNetV3-Small \cite{howard2019searching} & 2.94 & 0.07 & 54.351 & 84.343 & 74.426 & 51.807 & 66.232 & ICCV2019\\
    SwiftFormer-XS \cite{Shaker_2023_ICCV} &3.48 &0.62 & 58.099 & 87.789 & 79.649 & 62.651 & 72.047 & ICCV2023\\
    EfficientFormerv2-S0 \cite{li2022rethinking} & 3.60 & 0.40 & 53.055 &88.323 & 70.203 & 68.072 & 69.913 & ICCV2023\\
    FasterNet-T0 \cite{chen2023run} & 3.91 & 0.34 & 52.021 & 85.583 & 76.386 & 53.976 & 66.992 & CVPR2023\\
    FastViT-T8 \cite{vasufastvit2023} & 4.03 & 0.56 & 58.161 & 88.659 & 79.040 & 59.398 & 71.312 & ICCV2023\\
    EdgeViT-XXS \cite{pan2022edgevits} & 4.07 & 0.55 & 61.200 & 87.811 & 81.252 & 64.940 & 73.801 & ECCV2022\\
    Mobileone-S0 \cite{vasu2023mobileone} & 5.29 & 1.10 & 57.467 & 84.321 & 74.733 & 52.294 & 67.204 & CVPR2023\\
    MobileNetV3-Large \cite{howard2019searching} & 5.48 & 0.24 & 61.077 & 87.789 & 79.891 & 61.807 & 72.641 & ICCV2019\\
    
    \rowcolor{LightCyan}
    \textbf{Fraesormer-Tiny (Ours)} & 2.56 & 0.43 & \textbf{64.548} & \textbf{90.257} & \textbf{81.584} & \textbf{70.578} & \textbf{76.742} & \textbf{--} \\
    \midrule
    SwiftFormer-S \cite{Shaker_2023_ICCV} & 6.09 & 1.01 & 59.488 & 88.851 & 79.767 & 66.988 & 73.774 & ICCV2023\\
    GhostNetV2-1.0 \cite{tang2022ghostnetv2} & 6.16 & 0.18 & 58.917 & 87.981 & 80.144 & 56.265 & 70.827 & NIPS2022\\
    EMO-6M \cite{emo} & 6.17 & 0.95 & 63.206 & 89.002 & 80.144 & 68.554 & 72.227 & ICCV2023\\
    EfficientFormerV2-s1 \cite{li2022rethinking} & 6.19 & 0.67 & 55.955 & 89.877 & 73.693 & 68.916 & 72.110 & ICCV2023\\
    SHViT-S1 \cite{yun2024shvit} & 6.33 & 0.25 & 57.621 & 86.224 & 76.450 & 60.723 & 70.255 & CVPR2024\\
    EdgeViT-XS \cite{pan2022edgevits} & 6.77 & 1.13 & 62.465 & 89.810 & 82.762 & 70.120 & 76.289 & ECCV2022\\
    FastViT-T12 \cite{vasufastvit2023} & 7.55 & 1.11 & 59.873 & 89.471 & 81.015 & 64.096 & 73.614 & ICCV2023\\
    FasterNet-T1 \cite{chen2023run} & 7.60 & 0.86 & 53.193 & 87.383 & 78.708 & 56.386 & 68.918 & CVPR2023\\
    EfficientNet-B1 \cite{pmlr-v97-tan19a} & 7.79 & 0.60 & 65.659 & 90.386 & 82.480 & 69.036 & 76.890 & ICML2019\\
    MobileOne-S2 \cite{vasu2023mobileone} & 7.88 & 1.35 & 63.067 & 88.268 & 80.950 & 61.928 & 73.553 & CVPR2023\\
    MobileViTV2-1.3 \cite{mehta2022separable} & 8.10 & 2.42 & 63.345 & 89.412 & 81.520 & 69.012 & 75.822 & Arxiv2022\\
    GhostNetV2-1.3 \cite{tang2022ghostnetv2} & 8.96 & 0.29 & 60.753 & 89.028 & 81.163 & 59.639 & 72.646 & NIPS2022\\
    
    \rowcolor{LightCyan}
    \textbf{Fraesormer-Base (Ours)} & 6.39 & 1.21 & \textbf{68.050} & \textbf{90.799} & \textbf{83.054} & \textbf{76.024} & \textbf{79.482} & \textbf{--}\\
    \midrule
    SHViT-S2 \cite{yun2024shvit} & 11.48 & 0.37 & 57.282  & 87.176 & 76.584 & 61.259 & 70.575 & CVPR2024\\
    FastViT-Sa12 \cite{vasufastvit2023} & 11.58 & 1.52 & 62.789  & 89.523 & 81.322 & 64.819 & 74.613 & ICCV2023\\
    PoolFormer-S12 \cite{yu2022metaformer} & 11.92 & 1.83 & 52.823  & 85.996 & 77.490 & 61.264 & 69.393 & CVPR2022\\
    EfficientFormerV2-S2 \cite{li2022rethinking} & 12.71 & 1.27 &  50.741 & 88.504 & 59.797 & 70.120 & 67.291 & ICCV2023\\
    EdgeViT-S \cite{pan2022edgevits} & 13.11 & 1.91 & 61.031  & 88.401 & 82.624 & 68.313 & 75.092 & ECCV2022\\
    RepViT-M1-5 \cite{wang2024repvit} & 14.10 & 2.32 & 59.719  & 89.216 & 82.475 & 65.060 & 74.118 & CVPR2024\\
    SHViT-S4 \cite{yun2024shvit} & 16.59 & 0.79 & 60.830  & 88.475 & 79.490 & 64.458 & 73.313 & CVPR2024\\
    PoolFormer-S24 \cite{yu2022metaformer} & 21.39 & 3.42 & 59.904  & 89.589 & 79.520 & 63.256 & 73.067 & CVPR2022\\
    FasterNet-S \cite{chen2023run} & 31.18 & 4.57 &51.095  & 84.741 & 80.248 & 61.234 & 69.330 & CVPR2023\\
    
    \rowcolor{LightCyan}
    \textbf{Fraesormer-Large (Ours)}  & 10.39 & 1.74 & \textbf{67.957} & \textbf{91.515} & \textbf{85.460} & \textbf{75.542} & \textbf{80.119} & \textbf{--}\\
   
    \bottomrule[1.5pt]
    \end{tabular}
  }
  \end{center}
  \vspace{-5mm}
\end{table}
\begin{table}[!h]
\begin{center}
  \caption{Ablation experiments of different components.}
  \label{tab:ablation1}
  \vspace{-0.2cm}
  \resizebox{1\linewidth}{!}{
    \begin{tabular}{l|ccc}
    \toprule[0.15em]
     Model & \makecell{Params (M)$\downarrow$} & \makecell{MACs (G)$\downarrow$} & \makecell{Top-1 (\%)$\uparrow$}   \\
    \midrule[0.1em]
    w/o ADK-SPA  & 1.99 & 0.33 & 62.715 \\
    w/o HSSFGN & 1.52 & 0.24 & 60.077  \\
    ours  & 2.56 & 0.43 & 64.548\\
    \bottomrule[0.15em]
    \end{tabular}
  }
  \end{center}
  \vspace{-9mm}
\end{table}
where \(Split(\cdot)\) denotes channel-wise splitting, \( W(\cdot)_p \) is the \( 1 \times 1 \) point-wise convolution, \( W(\cdot)_d \) is the \( 3 \times 3 \) depth-wise convolution, and \( W(\cdot)_p^O \) denotes the projection weights for the output layer. \( \text{Concat}(\cdot) \) denotes the concatenation operation, while \( T_k(\cdot) \) represents the Gated Dynamic Top-K operator. It is important to note that, similar to traditional multi-head self-attention \cite{dosovitskiy2020vit}, we divide the channels into "heads" and compute separate attention maps in parallel. Additionally, we apply the \( 1 \times 1 \) point-wise convolution as the final projection layer across all channels, not just the initially divided attention channels. This ensures that the attention features extend to other channels while enhancing the local context, preserving information, and facilitating channel fusion and collaboration.

\vspace{-2.7mm}
{\flushleft\textbf{Gated Dynamic Top-k Operator (GDTKO)}.}
Since using a fixed \( k \) lacks the flexibility and dynamic adaptability needed to explore the potential of sparsity, we propose a Top-K operator that dynamically allocates and modulates based on the input. This allows the model to optimize attention distribution under both sparse and dense information conditions. The architecture of GDTKO is shown in Fig. \ref{overview}. Given the normalized feature \( F \in \mathbb{R}^{H \times W \times C'} \) from partial channels, we first use a Gated Unit to evaluate the contribution of the input feature map, capturing key information while suppressing noise. This produces a supervision modulation feature that retains the same spatial dimensions but contains only one channel. We then flatten this feature and compute its average value. Finally, we multiply the obtained supervised average by the channel dimension \( C' \) of the input feature \( F \). This ensures a proportional relationship between the value of \( k \) and the feature dimension, preventing imbalanced performance in low- or high-dimensional cases. Moreover, this enables the model to automatically adjust its behavior as the feature dimension changes, thereby improving its generalization ability.

\subsection{Hierarchical Scale-Sensitive Feature Gating Network}
The standard feed-forward network (FFN) usually processes information at each pixel independently, which is crucial for the self-attention mechanism \cite{dosovitskiy2020vit}. However, it is limited by single-scale feature extraction and overlooks the significance of multi-scale information in food recognition. Traditional multi-scale feed-forward networks tend to be bulky and redundant, requiring more computational resources. Therefore, we propose a more elegant approach to capture multi-scale local contextual information. Specifically, we design a scale-aware feed-forward network with a gate-like mechanism, incorporating depthwise separable convolutions at different scales (see Fig. \ref{overview}). On one hand, the gating mechanism alleviates the burden of processing redundant information, aligning with the design philosophy of channel attention by focusing on important features in the channel dimension. On the other hand, the proposed HSSFGN controls the flow of information across layers, allowing each layer to concentrate on the fine details necessary for the other layers, thus achieving cross-layer expert information fusion. Formally, HSSFGN can be expressed as
\vspace{-2mm}
\begin{equation}
    X_{G}=X_{V}=Linear(X_{l-1}),
    \vspace{-2mm}
\end{equation}
\begin{equation}
    X_{Gi}=Split(X_{G}), X_{Gi}'=f_{dw}^{i\times i}(X_{Gi}) ,i=[1,3,5,7],
    \vspace{-2mm}
\end{equation}
\begin{equation}
    \hat{X_{G}}=\delta (Concat(X_{Gi}')),
    \vspace{-2mm}
\end{equation}
\begin{equation}
    X_{l}=Linear(X_{V}\otimes \hat{X_{G}})+X_{l-1},
    \vspace{-2mm}
\end{equation}
where \( f_{dw}^{k \times k}(\cdot) \) represents the \( k \times k \) depth-wise convolution, \( \delta(\cdot) \) denotes the GELU activation function, \( \otimes \) indicates element-wise multiplication, and \( \text{Linear}(\cdot) \) represents the linear layer.

\section{Experiments and Analysis}

\subsection{Experimental Settings}

\vspace{-1mm}
{\flushleft\textbf{Datasets}.}
To evaluate the proposed model, we conducted experiments on four widely used food datasets: ETHZ Food-101 \cite{food-101}, Vireo Food-172 \cite{food-172}, UEC Food-256 \cite{food-256}, and SuShi-50 \cite{qiu2019mining}. 
Additionally, we divided the datasets into training, validation, and test sets in a 6:2:2 ratio.

\vspace{-2mm}
{\flushleft\textbf{Training Details}.}
We trained on images with a resolution of \(224 \times 224\), and all models were trained from scratch using the AdamW optimizer, without any pretraining or fine-tuning. The learning rate was set to \(1 \times 10^{-3}\), with a batch size of 256, for 300 epochs. We applied a cosine learning rate scheduler with a linear warm-up for the first 5 epochs. Data augmentation methods, including mixup augmentation and random erasing, were applied following the techniques described in \cite{yun2024shvit}. The entire framework was implemented in PyTorch and trained on an NVIDIA GeForce RTX 4090 GPU with 24GB of memory.

\subsection{Comparisons with The State-of-the-arts}
We compared our method with some open-source methods, including CNN-based approaches (RepViT \cite{wang2024repvit}, MobileNet v3 \cite{howard2019searching}, FasterNet \cite{chen2023run}, MobileOne \cite{vasu2023mobileone}, EfficientNet \cite{pmlr-v97-tan19a}, GhostNet v2 \cite{tang2022ghostnetv2}), ViT-based methods (SwiftFormer \cite{Shaker_2023_ICCV}, SHViT \cite{yun2024shvit}, PoolFormer \cite{yu2022metaformer}), and hybrid CNN-transformer approaches (EMO \cite{emo}, EfficientFormer v2 \cite{li2022rethinking}, FastViT \cite{vasufastvit2023}, EdgeViT \cite{pan2022edgevits}, MobileViT v2 \cite{mehta2022separable}).
The experimental results are shown in Table \ref{classification} and Fig. \ref{compare}. 
\begin{table}[!h]
\begin{center}
  \caption{Ablation studies of different self-attention.}
  \label{tab:ablation2}
  \vspace{-0.2cm}
  \resizebox{1\linewidth}{!}{
    \begin{tabular}{l|cccc}
    \toprule[0.15em]
     Model & \makecell{Params (M)$\downarrow$} & \makecell{MACs (G)$\downarrow$} & \makecell{Top-1 (\%)$\uparrow$} & \makecell{Year}  \\
    \midrule[0.1em]
    SDSA \cite{dosovitskiy2020vit}  & 2.95 & 0.48 & 62.892 & ICLR2021\\
    L-MHSA \cite{guo2022cmt} & 2.95 & 0.56 & 63.524  & CVPR2022\\
    H-MHSA \cite{liu2024vision} & 2.96 & 0.48 & 63.126  & MIR2024\\
    ATK-SPA  & 2.56 & 0.43 & 64.548 & -\\
    \bottomrule[0.15em]
    \end{tabular}
  }
  \end{center}
  \vspace{-0.5cm}
\end{table}
\begin{table}[!h]
\begin{center}
  \caption{Ablation studies of the PPCM.}
  \label{tab:ablation4}
  \vspace{-2mm}
  \resizebox{1\linewidth}{!}{
    \begin{tabular}{l|ccc}
    \toprule[0.15em]
     Model & \makecell{Params (M)$\downarrow$} & \makecell{MACs (G)$\downarrow$} & \makecell{Top-1 (\%)$\uparrow$}   \\
    \midrule[0.1em]
    w/o PPCM  & 3.11 & 0.51 & 64.559 \\
    w/ PPCM & 2.56 & 0.43 & 64.548  \\
    \midrule[0.1em]
    partial ratio = 1/8  & 2.21 & 0.39 & 62.921  \\
    partial ratio = 1/4 (ours) & 2.56 & 0.43 & 64.548 \\
    partial ratio = 1/2  & 2.89 & 0.48 & 64.556  \\
    \bottomrule[0.15em]
    \end{tabular}
  }
  \end{center}
  \vspace{-5mm}
\end{table}
\begin{table}[!h]
\begin{center}
  \caption{Ablation studies on alternatives to the HSSFGN.}
  \label{tab:ablation5}
  \vspace{-1.5mm}
  \resizebox{1\linewidth}{!}{
    \begin{tabular}{l|cccc}
    \toprule[0.15em]
     Model & \makecell{Params (M)$\downarrow$} & \makecell{MACs (G)$\downarrow$} & \makecell{Top-1 (\%)$\uparrow$} & \makecell{Year}  \\
    \midrule[0.1em]
    SFFN \cite{dosovitskiy2020vit}  & 2.97 & 0.48 & 62.805 & ICLR2021\\
    DFN \cite{li2021localvit}  & 5.44 & 0.59 & 64.982  & CVPR2021\\
    MixCFN \cite{gu2022multi} & 3.05 & 0.51 & 63.144  & CVPR2022\\
    ConvGLU \cite{shi2023transnext}& 3.00 & 0.49 & 62.959  & CVPR2024\\
    HSSFGN  & 2.56 & 0.43 & 64.548 & -\\
    \bottomrule[0.15em]
    \end{tabular}
  }
  \end{center}
  \vspace{-9mm}
\end{table}
We categorize the models into three groups based on parameter size. Within each parameter range, Fraesormer consistently outperforms other models in terms of Top-1 accuracy across all datasets while achieving a better balance between parameters and performance. Specifically, in the \(0 \text{--} 5M\) parameter range, Fraesormer-tiny achieves a Top-1 accuracy that is 6.83\% higher than that of EfficientFormerV2-S0 \cite{li2022rethinking}, while saving 29\% in parameters. It also surpasses MobileOne-S0 \cite{vasu2023mobileone} by 9.54\% in accuracy, while reducing parameters by 52\% and computational cost by 62\%. In the \(6 \text{--} 10M\) range, Fraesormer-Base's accuracy is 3.66\% higher than that of MobileViTV2-1.3 \cite{mehta2022separable}, with a 21\% reduction in parameters and a 50\% decrease in computational cost. In the \(> 10M\) parameter range, Fraesormer-Large achieves an accuracy that is 6.00\% higher than that of RepViT-M1-5 \cite{wang2024repvit}, while saving 26\% in parameters and 25\% in computational cost. Additionally, it outperforms FasterNet-S \cite{chen2023run} by 10.79\% in accuracy, saving 67\% in parameters and 62\% in computational cost. Furthermore, unlike previous Transformer models \cite{dosovitskiy2020vit}, Fraesormer excels not only on large datasets (ETHZ Food-101 \cite{food-101}) but also achieves optimal performance on smaller datasets (UEC-256 \cite{food-256} and SuShi-50 \cite{qiu2019mining}), indicating that Fraesormer can efficiently perform food recognition without requiring large data volumes or high computational costs.

\subsection{Ablation Studies}
{\flushleft\textbf{Ablation Experiments with Different Components}.}
To validate the effectiveness of various components in Fraesormer, we conducted ablation experiments on the UEC-256 dataset \cite{food-256} (Unless otherwise specified, all subsequent experiments use this dataset). The results are shown in Table \ref{tab:ablation1}. When we removed ATK-SPA, the model's accuracy decreased by 1.833\%. Similarly, removing HSSFGN led to a 4.471\% drop in accuracy. In Fraesormer, ATK-SPA adaptively extracts the most significant information in global contexts but cannot capture local contextual information. In contrast, HSSFGN focuses on capturing fine-grained local features across multiple scales and levels. Furthermore, HSSFGN and ATK-SPA complement each other, with HSSFGN suppressing redundancy in the channel dimension and ATK-SPA addressing redundancy in the spatial dimension. The collaboration of these two components enables Fraesormer to dynamically extract the most valuable insights at various levels while minimizing computational burden.

\begin{figure}[th]
\centering
    \includegraphics[width=0.35\textwidth]{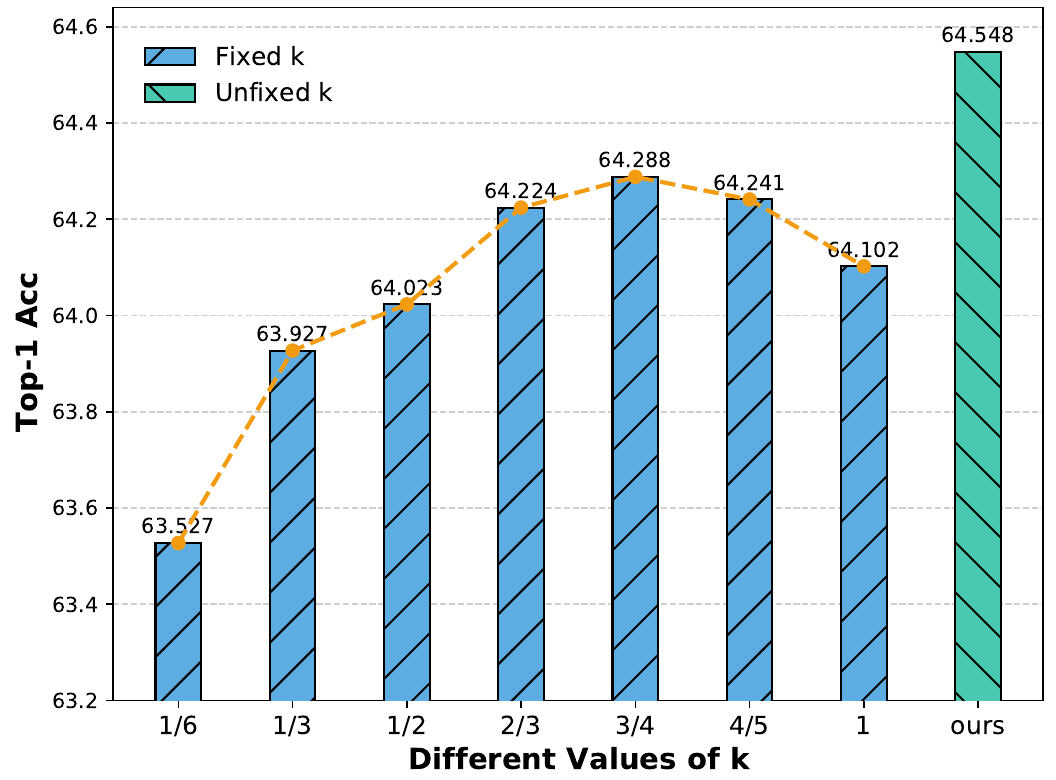}
  \vspace{-2mm}
  \caption{Ablation analysis of different values of $k$.}
  \label{fig:k}
  \vspace{-4mm}
\end{figure}
\begin{table}[h]
\begin{center}
  \caption{Ablation studies of the HSSFGN components.}
  \label{ablation_HSSFGN}
  \vspace{-1.5mm}
  \resizebox{1\linewidth}{!}{
  \begin{tabular}{cc|ccc}
    \toprule[0.15em]
     Gating Mechanism & Multi-scale Conv & Params (M)$\downarrow$ & MACs (G)$\downarrow$& Top-1 (\%)$\uparrow$ \\
    \midrule[0.1em]
    \XSolidBrush & \XSolidBrush & 2.52 & 0.41 & 63.406 \\
    \CheckmarkBold & \XSolidBrush & 2.52 & 0.41 & 64.231 \\
    \XSolidBrush & \CheckmarkBold & 2.56 & 0.43 &  64.143\\
    \CheckmarkBold & \CheckmarkBold & 2.56 & 0.43 & 64.548 \\
    \bottomrule[0.15em]
    \end{tabular}
    }
\end{center}
\vspace{-8mm}
\end{table}

\vspace{-2mm}
{\flushleft\textbf{Adaptive Top-k Sparse Partial Attention}.}
To validate the effectiveness of ATK-SPA, we replaced it with several recent efficient attention mechanisms: Standard Dense Self-Attention (SDSA) \cite{dosovitskiy2020vit}, Lightweight Multi-Head Self-Attention (L-MHSA) \cite{guo2022cmt}, and Hierarchical Multi-Head Self-Attention (H-MHSA) \cite{liu2024vision}. The quantitative results are shown in Table \ref{tab:ablation2}. Compared to SDSA, L-MHSA, and H-MHSA, our ATK-SPA demonstrated superior performance in both accuracy and parameter efficiency.

\vspace{-2mm}
{\flushleft\textbf{Gated Dynamic Top-k Operator}.}
The key factor for GDTKO is the value of \( k \). To verify the effectiveness of the adaptive \( k \) value, we fixed different \( k \) values and compared them with the dynamic \( k \) value from GDTKO, as shown in Fig. \ref{fig:k}. We observed that setting \( k \) to a smaller value led to a significant loss of useful information. Conversely, selecting a \( k \) value that is too large, or even including all tokens for attention computation, not only increased the computational burden but also introduced interference from irrelevant tokens. Thus, manually setting the \( k \) value fails to adaptively adjust the network based on input data, while GDTKO effectively balances sparsity and density to dynamically capture the most influential tokens.

\vspace{-2mm}
{\flushleft\textbf{Parallel Partial Channel Mechanism}.}
To investigate the effectiveness of the partial channel mechanism, we removed this mechanism, and the experimental results are shown in Table \ref{tab:ablation4}. When the partial channel mechanism is applied, we achieve comparable accuracy with lower parameters and computational overhead. This demonstrates that the method enables token interactions across channels at a reduced cost, achieving a high cost-performance ratio. 
Furthermore, we explored the optimal ratio configuration. When the partial ratio is set to 1/4 (the default setting of Fraesormer), it achieves the best trade-off between accuracy and speed. Compared to very small values, a moderate increase in token interactions across channels effectively improves performance at a low cost. However, excessively large values fail to provide a cost-effective performance improvement, as the accompanying computational cost outweighs the benefits.

\vspace{-2mm}
{\flushleft\textbf{Hierarchical Scale-Sensitive Feature Gating Network}.}
To demonstrate the effectiveness of HSSFGN, we compared it with four variants: Standard Feed-Forward Network (SFFN) \cite{dosovitskiy2020vit}, Depth-wise Convolution Equipped Feed-forward Network (DFN) \cite{li2021localvit}, Mixed-scale Convolutional Feedforward Network (MixCFN) \cite{gu2022multi}, and Convolutional Gated Linear Unit (ConvGLU) \cite{shi2023transnext}. The quantitative comparison is presented in Table \ref{tab:ablation5}. Compared to SFFN, MixCFN, and ConvGLU, our HSSFGN achieved the best performance in terms of accuracy, parameter efficiency, and computational cost. Although DFN's accuracy is slightly higher than that of HSSFGN by 0.43\%, it incurs an additional 53\% in parameters and 27\% in computational cost. Furthermore, we further explore the components of HSSFGN in Table \ref{ablation_HSSFGN}. The results demonstrate that both the gated mechanism paradigm and the multi-scale convolution in the gated branch are indispensable to HSSFGN. The gated mechanism paradigm effectively promotes information flow, preventing blockage and redundancy. Meanwhile, the multi-scale convolution in the gated branch extracts multi-scale expert information across channels, working synergistically with the gated mechanism to enhance feature representation.

\section{Conclusion}
This paper introduces an adaptive efficient sparse Transformer for food recognition, named Fraesormer. To adaptively model global long-range dependencies while enhancing the sparsity of the neural network, reducing redundant information, and minimizing computational burden, we developed an Adaptive Top-k Sparse Partial Attention to replace standard self-attention. Furthermore, we implemented an efficient multi-scale feedforward network based on a gating mechanism to further improve feature representation capabilities. Extensive experimental results indicate that Fraesormer achieves a favorable balance between performance and computational efficiency.

\bibliographystyle{IEEEbib}
\bibliography{references-main}

\end{document}